\documentclass[11pt,a4paper]{article}
\usepackage[hyperref]{emnlp2018}
\usepackage{times}
\usepackage{latexsym}

\usepackage{graphicx}
\usepackage{amsmath,amsthm,amssymb,amsfonts}
\usepackage{tabularx}
\usepackage{multirow}
\usepackage{makecell}
\usepackage{array}
\usepackage{tablefootnote}

\usepackage[]{url}

\aclfinalcopy % Uncomment this line for the final submission
\title{Sequence Labeling: A Practical Approach}

\author{Adnan Akhundov \hspace{0.5cm} Dietrich Trautmann \hspace{0.5cm} Georg Groh \\
  Department of Informatics, Technical University of Munich \\
  {\tt \{adnan.akhundov, grohg\}@in.tum.de} \\
  {\tt dietrich@trautmann.me}}

\date{}

\begin{document}

\maketitle

\begin{abstract}
  We take a practical approach to solving sequence labeling problem assuming unavailability of domain expertise and scarcity of informational and computational resources. To this end, we utilize a universal end-to-end Bi-LSTM-based neural sequence labeling model applicable to a wide range of NLP tasks and languages. The model combines morphological, semantic, and structural cues extracted from data to arrive at informed predictions. The model's performance is evaluated on eight benchmark datasets (covering three tasks: POS-tagging, NER, and Chunking, and four languages: English, German, Dutch, and Spanish). We observe state-of-the-art results on four of them: \mbox{CoNLL-2012} (English NER), \mbox{CoNLL-2002} (Dutch NER), \mbox{GermEval 2014} (German NER), Tiger Corpus (German POS-tag.), and competitive performance on the rest. Our source code and detailed experimental results are publicly available\footnote{\url{https://github.com/aakhundov/sequence-labeling}}.
\end{abstract}

\section{Introduction}

A variety of NLP tasks can be formulated as general sequence labeling problem: given a sequence of tokens and a fixed set of labels, assign one of the labels to each token in a sequence. We consider three concrete sequence labeling tasks: Part-of-speech (POS) tagging, Named Entity Recognition (NER), and Chunking (also known as shallow parsing). POS-tagging reduces to assigning a part-of-speech label to each word in a sentence; NER requires detecting (potentially multi-word) named entities, like person or organization names; chunking aims at identifying syntactic constituents within a sentence, like noun- or verb-phrases.

Traditionally, sequence labeling tasks were tackled using linear statistical models, for instance: Hidden Markov Models \cite{kupiec1992robust}, Maximum Entropy Markov Models \cite{mccallum2000maximum}, and Conditional Random Fields \cite{lafferty2001conditional}. In their seminal paper, \newcite{collobert2011natural} have introduced a deep neural network-based solution to the problem, which has spawned immense research in this direction. Multiple works have introduced different neural architectures for universal sequence labeling afterwards \cite{huang2015bidirectional, lample2016neural, ma2016end, chiu2016named, yang2016multi}. However, aiming for better results on a particular dataset, these and numerous other works typically employ some form of feature engineering \cite{ando2005framework, shen2005voting, collobert2011natural, huang2015bidirectional},  external data for training \cite{ling2015finding, lample2016neural} or in a form of lexicons and gazetteers \cite{ratinov2009design, passos2014lexicon, chiu2016named}, extensive hyper-parameter search \cite{chiu2016named, ma2016end}, or multi-task learning \cite{durrett2014joint, yang2016multi}.

\begin{figure*}[t]
    \small
    \fontfamily{cmss}\selectfont
    \includegraphics[width=\textwidth]{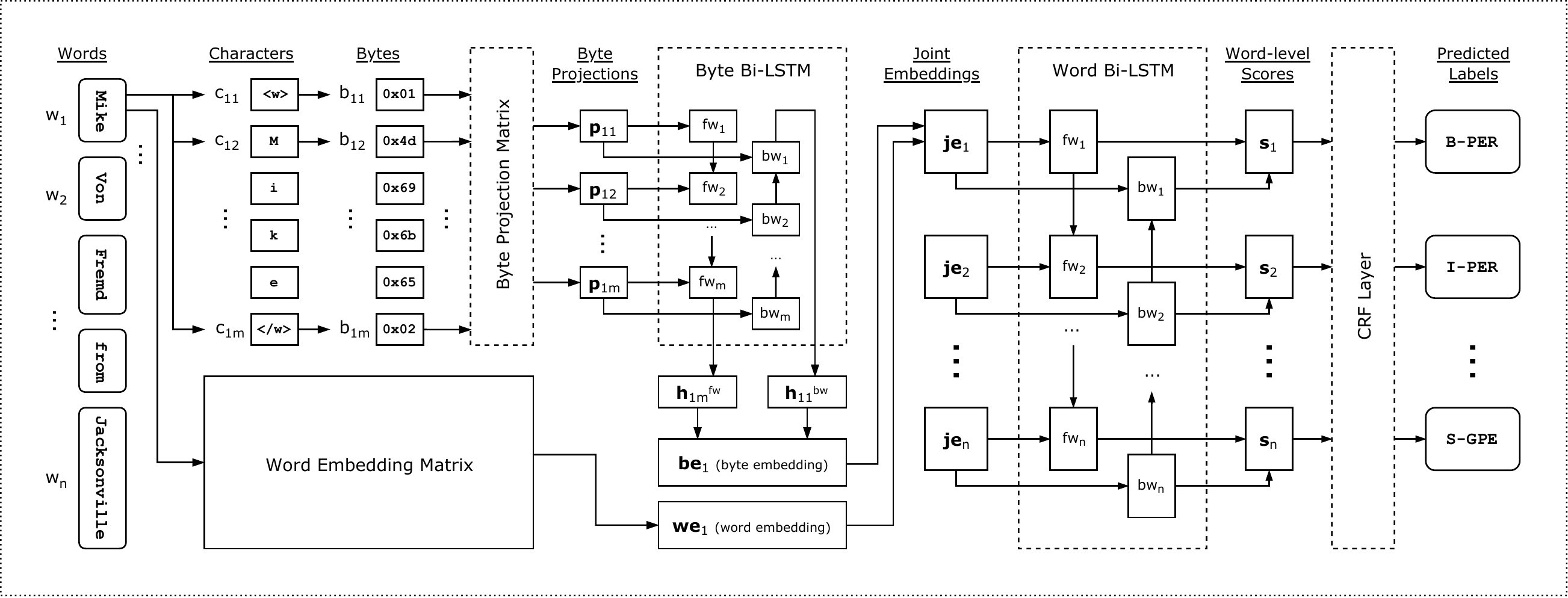}
    \centering
    \caption{Diagram of the proposed sequence labeling model. Learned components are shown in dashed rectangles. For clarity, computing byte and word embeddings is shown only for the first word of the sentence.}
    \label{diagram}
\end{figure*}

In this paper, we take an alternative stance by looking at the problem of sequence labeling from a practical perspective. The performance enhancements enumerated above typically require availability of expertise, time, or external resources. Some (or even all) of these may be unavailable to users in a practical situation. Therefore, we deliberately eschew any form of feature engineering, pre-training, external data (with the exception of publicly available word embeddings), and time-consuming hyper-parameter optimization. To this end, we formulate a single general-purpose sequence labeling model and apply it to eight different benchmark datasets to estimate the effectiveness of our approach.

Our model utilizes a bi-directional LSTM \cite{graves2005framewise} to extract morphological information from the bytes of words in a sentence. These, combined with word embeddings bearing semantic cues, are fed to another Bi-LSTM to obtain word-level scores. Ultimately, the word-level scores are passed through a CRF layer \cite{lafferty2001conditional} to facilitate structured prediction of the labels.

The rest of the paper is organized as follows. Section \ref{model} specifies the proposed model in detail. Sections \ref{datasets} and \ref{training} describe the datasets and the training procedure used in experiments. The results are presented in Section \ref{results}. We review related work in Section \ref{related} and conclude in Section \ref{conclusion}.

\section{Model} \label{model}

Recurrent Neural Networks \cite{elman1990finding} are commonly used for modeling sequences in NLP \cite{mikolov2010recurrent, cho2014properties, graves2013speech}. However, because of well-known challenges of capturing long-term dependencies with plain RNNs \cite{pascanu2013difficulty}, we instead turn to Long Short-Term Memory (LSTM) networks \cite{hochreiter1997long} capable of alleviating the vanishing/exploding gradient problem by design. The specific LSTM formulation we are using \cite{zaremba2014recurrent} can be described by the following equations:
\begin{align*}
\textbf{i}_{t} & = \sigma(\textbf{W}_{i} \textbf{h}_{t-1} + \textbf{U}_{i} \textbf{x}_{t} + \textbf{b}_{i}) \\
\textbf{f}_{t} & = \sigma(\textbf{W}_{f} \textbf{h}_{t-1} + \textbf{U}_{f} \textbf{x}_{t} + \textbf{b}_{f}) \\
\textbf{o}_{t} & = \sigma(\textbf{W}_{o} \textbf{h}_{t-1} + \textbf{U}_{o} \textbf{x}_{t} + \textbf{b}_{o}) \\
\widetilde{\textbf{c}_{t}} & = tanh(\textbf{W}_{c} \textbf{h}_{t-1} + \textbf{U}_{c} \textbf{x}_{t} + \textbf{b}_{c}) \\
\textbf{c}_{t} & = \textbf{f}_{t} \odot \textbf{c}_{t-1} + \textbf{i}_{t} \odot \widetilde{\textbf{c}_{t}} \\
\textbf{h}_{t} & = \textbf{o}_{t} \odot tanh(\textbf{c}_{t})
\end{align*}
where $\sigma$ denotes the sigmoid activation function; $\odot$ - element-wise (Hadamard) product; $\textbf{W}, \textbf{U}, \textbf{b}$ - learned network parameters; $\textbf{i}_{t}, \textbf{f}_{t}, \textbf{o}_{t}$ - input, forget, and output gates; $\textbf{x}_{t}$, $\textbf{c}_{t}$, and $\textbf{h}_{t}$ - network input, cell state, and network output at time step $t$ respectively.

One issue with ordinary LSTMs is that they capture dependencies between sequence elements only in one direction, whereas it might be beneficial to learn also backward dependencies (e.g., for informed first label prediction). To overcome this limitation, we use bidirectional LSTM (Bi-LSTM) networks \cite{graves2005framewise} comprising two independent LSTM instances (with separate sets of parameters): one processing an input sequence in forward direction and the other in backward direction. The output of Bi-LSTM is formed by concatenating the outputs of the two LSTMs corresponding to each sequence element.

The diagram of the proposed model is shown in Figure \ref{diagram}. The model can be decomposed into three logical components: (1) computing byte embeddings of each word in a sentence by means of the Byte Bi-LSTM; (2) combining byte embeddings with word embeddings and feeding resulting joint word representations to the Word Bi-LSTM to obtain word-level scores; (3) passing word-level scores through the CRF Layer to arrive at joint prediction of labels. We describe each of these components in detail in the following subsections.

\subsection{Byte Embeddings}

Assuming that input is available in tokenized form, we enable the model to extract morphological information from tokens by analyzing their character-level representations. Following \newcite{ling2015finding}, we apply a Bi-LSTM network for solving this task. However, to be truly neutral with respect to languages and character sets thereof, we consider a sequence of bytes underlying a \mbox{UTF-8} encoding of each word instead of its characters.

Formally, given a sequence of words $w_{1}, ..., w_{n}$ in a sentence, we first decompose each word into its characters $c_{i1}, ..., c_{im_{i}}$ including dummy start- and end-of-word characters (we assume that the \mbox{$i$-th} word consists of $m_i$ characters including two dummy ones). Now we convert a sequence of characters to a sequence of underlying \mbox{UTF-8} bytes $b_{i1}, ..., b_{im_{i}}$ (for simplicity, here we assume that all characters are single-byte, but this is obviously not necessary). Dummy characters are encoded by special byte values $\texttt{0x01}$ (start-of-word) and $\texttt{0x02}$ (end-of-word). Next, we compute byte projections of bytes in a sequence by multiplying the byte projection matrix $\textbf{B}$ by a one-of-$256$ coded vector representation of each byte. The matrix $\textbf{B} \in \mathbb{R}^{d_{b} \times 256}$ is a learned model parameter. Each of the obtained byte projection vectors $\textbf{p}_{i1}, ..., \textbf{p}_{im_{i}}$ has a fixed dimensionality $d_{b}$, which is a hyper-parameter of the model.

Next, the byte projection sequence $\textbf{p}_{i1}, ..., \textbf{p}_{im_{i}}$ is fed to the Byte Bi-LSTM network. The last outputs of its forward and backward LSTMs are concatenated to obtain "byte embedding" vector of the \mbox{$i$-th} word $\textbf{be}_{i}$. These fixed-size vectors are assumed to capture morphological information about corresponding words in a sentence.

\subsection{Word-level Representation}

Morphological information alone is probably not representative enough to reliably predict target labels in a general sequence labeling setting. For this reason, we would also want to supply our prediction framework with semantic information. We fulfill this requirement by mixing in pre-trained word embeddings of words in a sentence. Then, following \newcite{huang2015bidirectional}, we infer word-level representation using a Bi-LSTM network.

Formally, given a vocabulary of size $V$ and fixed (not learned) word embedding matrix $\textbf{E} \in \mathbb{R}^{d_{e} \times V}$, the word embedding vector of the \mbox{$i$-th} word $\textbf{we}_{i}$ is obtained by multiplying $\textbf{E}$ by a one-of-$V$ coded vector representation of the word's position in the vocabulary ($d_{e}$ is the dimensionality of embedding vectors and depends on the choice of word embeddings). For all out-of-vocabulary words we use the same additional "unknown" word embedding vector.

Next, computed byte embedding of \mbox{$i$-th} word $\textbf{be}_{i}$ is concatenated with its word embedding $\textbf{we}_{i}$ to produce a joint embedding $\textbf{je}_{i}$. This way we obtain a sequence of joint embeddings $\textbf{je}_{1}, ..., \textbf{je}_{n}$ corresponding to the words $w_{1}, ..., w_{n}$ in a sentence. The joint embeddings are assumed to capture both morphological and semantic information about the words. They are fed as inputs to the Word Bi-LSTM network. The outputs of the network at each time step are passed through a linear layer (with no activation function) to yield $L$-dimensional word-level score vectors $\textbf{s}_{1}, ..., \textbf{s}_{n}$, where $L$ denotes the number of distinct labels. In essence, word-level score vectors may be interpreted as unnormalized log-probabilities (logits) over the labels at each time step.

Theoretically, we could stop here by applying softmax to each word-level score vector to infer a distribution over possible labels at each time step. However, this approach (albeit bearing a certain degree of context-awareness due to the presence of the upstream Word Bi-LSTM) would treat each word more or less locally, lacking a global view over predicted labels. This is why we turn to a CRF layer as the last step of label inference.

\begin{table*}[t]
    \small
    \fontfamily{cmss}\selectfont
    \begin{tabularx}{\textwidth}{|X|>{\centering\arraybackslash}m{3em}>{\centering\arraybackslash}m{3.9em}|>{\centering\arraybackslash}m{3em}>{\centering\arraybackslash}m{3.9em}|>{\centering\arraybackslash}m{3em}>{\centering\arraybackslash}m{3.9em}|>{\centering\arraybackslash}m{3.5em}|@{}m{0pt}@{}}
        \hline
        \multirow{2}{*}{\parbox{1\linewidth}{\vspace{0.1cm} \textbf{Dataset}}} & \multicolumn{2}{c|}{\textbf{training set}} & \multicolumn{2}{c|}{\textbf{development set}} & \multicolumn{2}{c|}{\textbf{testing set}} &
        \multirow{2}{*}{\parbox{1\linewidth}{\vspace{0.05cm} \makecell{\textbf{labels} \\ (\textbf{classes})}}} & \\ [1.0ex]
        \cline{2-7}
        & \textbf{sentences} & \textbf{tokens} & \textbf{sentences} & \textbf{tokens} & \textbf{sentences} & \textbf{tokens} & & \\ [1.0ex]
        \hline
        \textbf{CoNLL 2000} (English, Chunking) & 8,936 & 211,727 & - & - & 2,012 & 47,377 & 45 (11) & \\ [0.5ex]
        \hline
        \textbf{CoNLL 2002} (Spanish, NER) & 8,323 & 264,715 & 1,915 & 52,923 & 1,517 & 51,533 & 17 (4) & \\ [0.5ex]
        \hline
        \textbf{CoNLL 2002} (Dutch, NER) & 15,806 & 202,644 & 2,895 & 37,687 & 5,195 & 68,875 & 17 (4) & \\ [0.5ex]
        \hline
        \textbf{CoNLL 2003} (English, NER) & 14,041 & 203,621 & 3,250 & 51,362 & 3,453 & 46,435 & 17 (4) & \\ [0.5ex]
        \hline
        \textbf{CoNLL 2012} (English, NER) & 59,924 & 1,088,503 & 8,528 & 147,724 & 8,262 & 152,728 & 73 (18) & \\ [0.5ex]
        \hline
        \textbf{GermEval 2014} (German, NER) & 24,000 & 452,853 & 2,200 & 41,653 & 5,100 & 96,499 & 49 (12) & \\ [0.5ex]
        \hline
        \textbf{WSJ/PTB} (English, POS-tag.) & 38,219 & 912,344 & 5,527 & 131,768 & 5,462 & 129,654 & 45 (45) & \\ [0.5ex]
        \hline
        \textbf{Tiger Corpus} (German, POS-tag.) & 40,472 & 719,530 & 5,000 & 76,704 & 5,000 & 92,004 & 54 (54) & \\ [0.5ex]
        \hline
    \end{tabularx}
    \centering
    \caption{Statistics of eight benchmark datasets used in the experiments.}
    \label{table_datasets}
\end{table*}

\subsection{CRF Layer} \label{crf_layer}

Oftentimes, labels predicted at different time steps follow certain structural patterns. As an example, the IOB labeling scheme has specific rules constraining label transitions: for example, an \mbox{I-ORG} label can follow only a \mbox{B-ORG} or another \mbox{I-ORG} but no other label. To learn patterns like this one, following \newcite{lample2016neural}, we utilize Conditional Random Fields (CRFs) \cite{lafferty2001conditional} in the final component of our model. CRFs can capture dependencies between labels predicted at different time steps by modeling probabilities of transitions from one step to the other. Linear chain CRFs model transitions between neighboring pairs of labels in a sequence and allow solving a structured prediction problem in a computationally feasible way.

We recall that a sequence of word-level score vectors $\textbf{s}_{1}, ..., \textbf{s}_{n}$ is inferred by the Word Bi-LSTM. The \mbox{$j$-th} component of the \mbox{$i$-th} score vector - $s_{ij}$ - represents unnormalized log-probability of assigning \mbox{$j$-th} label to the word at the \mbox{$i$-th} position. Joint prediction is modeled by introducing a total score $f$ of a sequence of labels $\textbf{y} = y_{1}, ..., y_{n}$ given a sequence of words $\textbf{w} = w_{1}, ..., w_{n}$:
\begin{align*}
f(\textbf{y}|\textbf{w}) = \sum_{i=1}^{n} s_{iy_{i}} + \sum_{i=1}^{n-1} A_{y_{i}y_{i+1}}
\end{align*}
where $\textbf{A} \in \mathbb{R}^{L \times L}$ is a matrix of label transition scores ($A_{ij}$ is a score of transition from label $i$ to label $j$; $L$ represents the number of distinct labels) and word-level scores $s_{iy_{i}}$ obviously depend on $\textbf{w}$. The matrix $\textbf{A}$ is another learned model parameter.

The probability of observing a particular sequence of labels $\textbf{y} = y_{1}, ..., y_{n}$ given a sequence of words $\textbf{w} = w_{1}, ..., w_{n}$ can be computed by applying softmax over total scores of all possible label assignments $\widetilde{\textbf{y}}$ to a sequence of words $\textbf{w}$ ($\theta$ denotes the set of all learned model parameters):
\begin{align} \label{prob}
p(\textbf{y}|\textbf{w};\theta) = \frac{e^{f(\textbf{y}|\textbf{w})}}{\sum_{\widetilde{\textbf{y}}} e^{f(\widetilde{\textbf{y}}|\textbf{w})}}
\end{align}
And the corresponding log-probability is:
\begin{align} \label{logprob}
\log p(\textbf{y}|\textbf{w};\theta) = f(\textbf{y}|\textbf{w}) - \log \sum_{\widetilde{\textbf{y}}} e^{f(\widetilde{\textbf{y}}|\textbf{w})}
\end{align}

We learn the model parameters by maximizing log-likelihood (\ref{logprob}) of $\theta$ given the ground truth labels \textbf{y} corresponding to the input sequence \textbf{w}. Computing the normalizing factor from \mbox{equation (\ref{prob})}, as well as deriving the most likely sequence of labels during test time is performed using dynamic programming \cite{rabiner1989tutorial}.

\section{Datasets} \label{datasets}

We evaluate the performance of our approach on eight benchmark datasets covering four languages and three sequence labeling tasks. Certain statistics of those are shown in Table \ref{table_datasets}. The labels of all NER and Chunking datasets are converted to IOBES tagging scheme, as it is reported to increase predictive performance \cite{ratinov2009design}. We briefly discuss each of the datasets in the subsections below.

\subsection{CoNLL 2000}

The CoNLL 2000 dataset \cite{tjong2000introduction} was introduced as a part of a shared task on Chunking. Sections 15-18 of the Wall Street Journal part of the Penn Treebank corpus \cite{marcus1993building} are used for training, section 20 for testing. Due to the lack of specifically demarcated development set, we use randomly sampled 10\% of the training set for this purposes (see Section \ref{training} for the details of training).

\subsection{CoNLL 2002}

The CoNLL 2002 dataset \cite{tjong2002introduction} was used for shared task on language-independent Named Entity Recognition. The data represents news wire covering two languages: Spanish and Dutch. In our experiments we treat Spanish and Dutch data separately, as two different datasets. The dataset is annotated by four entity types: persons (PER), organizations (ORG), locations (LOC), and miscellaneous names (MISC).

\subsection{CoNLL 2003}

CoNLL 2003 \cite{tjong2003introduction} is a NER dataset structurally similar to CoNLL 2002 (including entity types), but in English and German. English data is based on news stories from Reuters Corpus \cite{rose2002reuters}. We use the English portion of the dataset in the experiments.

\subsection{CoNLL 2012}

The CoNLL 2012 dataset \cite{pradhan2012conll} was created for a shared task on multilingual unrestricted coreference resolution. The dataset is based on OntoNotes corpus v5.0 \cite{hovy2006ontonotes} and, among others, has named entity annotations. It is substantially larger and more diverse than the previously described NER datasets (see Table \ref{table_datasets} for detailed comparison). Although some sources \cite{durrett2014joint, chiu2016named} refer to the dataset as "OntoNotes", we stick to the name "CoNLL 2012" as the train/dev/test split that is used by this and other works is not defined in the OntoNotes corpus. Following \newcite{durrett2014joint}, we exclude the New Testament part of the data, as it is lacking gold annotations.

\subsection{GermEval 2014}

GermEval 2014 \cite{benikova2014germeval} is a recently organized shared task on German Named Entity Recognition. The corresponding dataset has four main entity types (Location, Person, Organization, and Other) and two sub-types of each type, \mbox{"-deriv"} and \mbox{"-part"}, indicating derivation from and inclusion of a named entity respectively. The original dataset has two levels of labeling: outer and inner. However, we use only outer labels in the experiments and compare our results to other works on the "M3: Outer Chunks" metric.

\subsection{Wall Street Journal / Penn Treebank}

The Wall Street Journal section of the Penn Treebank corpus \cite{marcus1993building} is commonly used as a benchmark dataset for the English POS-tagging task. We follow this tradition and use a standard split of sections: 0-18 for training, 19-21 for development, and 22-24 for testing \cite{toutanova2003feature}.

\subsection{Tiger Corpus}

Tiger Corpus \cite{brants2002tiger} is an extensive collection of German newspaper texts. The dataset has several different types of annotations. We use part-of-speech annotations for setting up a German POS tagging task. Following \newcite{fraser2013knowledge}, we use the first 40,472 of the originally ordered sentences for training, the next 5,000 for development, and the last 5,000 for testing.

\section{Training} \label{training}

The training procedure described in this section is used for every experiment on every dataset mentioned in this paper. The model is trained end-to-end, accepting tokenized sentences as input and predicting per-token labels as output.

The model is trained using an Adam optimizer \cite{kingma2014adam}. Following \newcite{ma2016end}, we apply staircase learning rate decay:
\begin{align*}
\eta_{t} = \frac{\eta_{0}}{1 + \rho (t-1)}
\end{align*}
where $\eta_{t}$ is the learning rate used throughout \mbox{$t$-th} epoch ($t$ starts at $1$), $\eta_{0}$ is the initial learning rate, and $\rho$ is the learning rate decay factor. In our experiments we use $\eta_{0} = 10^{-3}$ and $\rho = 0.05$.

Motivated by the initial experiments, the dimension $d_{b}$ of the byte projections is set to $50$, the size of Byte and Word Bi-LSTM to $64$ and $128$ respectively. Training lasts for $100$ epochs with a batch size of $8$ sentences. Early stopping \cite{caruana2001overfitting} is used: at the end of every epoch the model is evaluated against the development set; eventually, the parameter values performing best on the development set are declared the final values.

Due to the high level of expressive power of the proposed model, we use dropout \cite{srivastava2014dropout}, with the rate of $0.5$, to reduce the possibility of overfitting. Dropout is applied to word embeddings and byte projections, as well as the outputs of Byte Bi-LSTM and Word Bi-LSTM. As per \newcite{zaremba2014recurrent}, we don't apply dropout to state transitions of the LSTM networks.

Publicly available word embeddings are used for every language in the experiments. For English datasets we use $100$-dimensional uncased GloVe embeddings \cite{pennington2014glove} trained on English Wikipedia and Gigaword 5 corpora and comprising $400$K unique word forms. For other languages we use $64$-dimensional cased Polyglot embeddings \cite{al2013polyglot} trained on a respective Wikipedia corpus and comprising $100$K unique (case-sensitive) word forms per language. Maintaining our commitment to the practical approach, we freeze the word embeddings during training not allowing them to train (except for the "unknown" word embedding, which is trained).

To achieve higher efficiency, we compute the joint embedding of every unique word in a batch only once \cite{ling2015finding}. Unique joint embeddings are scattered according to the word positions in the input sentences, before being fed to the Word Bi-LSTM. The gradient with respect to each unique byte embedding is accumulated before being back-propagated once through the Byte Bi-LSTM. Albeit marginal during training (in small batches), performance improvement becomes considerable during inference (in larger batches).

\begin{table*}[t]
    \small
    \fontfamily{cmss}\selectfont
    \begin{tabularx}{\textwidth}{|X|>{\centering\arraybackslash}m{4em}|>{\centering\arraybackslash}m{4em}|>{\centering\arraybackslash}m{4em}|>{\centering\arraybackslash}m{4em}|>{\centering\arraybackslash}m{4em}|>{\centering\arraybackslash}m{4em}|@{}m{0pt}@{}}
        \hline
        \textbf{Dataset} & \textbf{word} & \makecell{\textbf{word} \texttt{+} \\ \textbf{crf}} & \textbf{byte} & \makecell{\textbf{byte} \texttt{+} \\ \textbf{crf}} & \makecell{\textbf{byte} \texttt{+} \\ \textbf{word}} & \makecell{\textbf{byte} \texttt{+} \\ \textbf{word} \texttt{+} \\ \textbf{crf}} & \\ [6ex]
        \hline
        \textbf{CoNLL 2000} (English, Chunking, F$_{1}$) & 91.39 & 92.70 & 92.75 & 93.41 & 93.93 & \textbf{94.74} & \\ [0.5ex]
        \hline
        \textbf{CoNLL 2002} (Spanish, NER, F$_{1}$) & 77.55 & 80.31 & 77.29 & 79.62 & 82.04 & \textbf{84.36} & \\ [0.5ex]
        \hline
        \textbf{CoNLL 2002} (Dutch, NER, F$_{1}$) & 74.95 & 76.91 & 75.14 & 78.06 & 83.26 & \textbf{85.61} & \\ [0.5ex]
        \hline
        \textbf{CoNLL 2003} (English, NER, F$_{1}$) & 86.35 & 87.58 & 81.80 & 82.85 & 89.70 & \textbf{91.11} & \\ [0.5ex]
        \hline
        \textbf{CoNLL 2012} (English, NER, F$_{1}$) & 79.00 & 82.95 & 79.93 & 83.70 & 84.69 & \textbf{87.84} & \\ [0.5ex]
        \hline
        \textbf{GermEval 2014} (German, NER, F$_{1}$) & 68.86 & 71.90 & 65.79 & 69.30 & 76.74 & \textbf{79.21} & \\ [0.5ex]
        \hline
        \textbf{WSJ/PTB} (English, POS-tag., \%) & 95.22 & 95.30 & 97.11 & 97.14 & \textbf{97.45} & 97.43 & \\ [0.5ex]
        \hline
        \textbf{Tiger Corpus} (German, POS-tag., \%) & 95.60 & 95.76 & 97.68 & 97.78 & 98.39 & \textbf{98.40} & \\ [0.5ex]
        \hline
    \end{tabularx}
    \centering
    \caption{Results of the ablation studies. \textbf{word} and \textbf{byte} indicate inclusion of respective embeddings into joint embeddings; \textbf{crf} indicates presence of the CRF layer. Testing score of the best model is marked in bold.}
    \label{table_results_ablation}
\end{table*}

\begin{table}[ht]
    \small
    \vspace{0.2cm}
    \fontfamily{cmss}\selectfont
    \begin{tabular}{p{4.2cm}|>{\centering\arraybackslash}p{1.2cm}}
        Model & F$_{1}$ Score \\
        \hline
        \newcite{shen2005voting} & 94.01 \\
        \newcite{collobert2011natural} & 94.32 \\
        \newcite{sun2008modeling} & 94.34 \\
        \newcite{huang2015bidirectional} & 94.46 \\
        \newcite{zhai2017neural} & 94.72 \\
        \textbf{Our Model} & \textbf{94.74} \\
        \newcite{yang2016multi} & 95.41 \\
        \newcite{peters2017semi} & 96.37
    \end{tabular}
    \centering
    \parbox{6.2cm}{\caption{Comparison with other works on CoNLL 2000 dataset (English, Chunking).} \label{table_results_conll2000}}
\end{table}

\begin{table}[ht]
    \small
    \vspace{0.5cm}
    \fontfamily{cmss}\selectfont
    \begin{tabular}{p{4.2cm}|>{\centering\arraybackslash}p{1.2cm}}
        Model & F$_{1}$ Score \\
        \hline
        \newcite{carreras2002named} & 81.39 \\
        \newcite{dos2015boosting} & 82.21 \\
        \newcite{gillick2016multilingual} & 82.95 \\
        \textbf{Our Model} & \textbf{84.36} \\
        \newcite{lample2016neural} & 85.75 \\
        \newcite{yang2016multi} & 85.77 \\
    \end{tabular}
    \centering
    \parbox{6.2cm}{\caption{Comparison with other works on CoNLL 2002 dataset (Spanish, NER).} \label{table_results_conll2002_esp}}
\end{table}

\begin{table}[ht]
    \small
    \vspace{0.2cm}
    \fontfamily{cmss}\selectfont
    \begin{tabular}{p{4.2cm}|>{\centering\arraybackslash}p{1.2cm}}
        Model & F$_{1}$ Score \\
        \hline
        \newcite{carreras2002named} & 77.05 \\
        \newcite{nothman2013learning} & 78.60 \\
        \newcite{lample2016neural} & 81.74 \\
        \newcite{gillick2016multilingual} & 82.84 \\
        \newcite{yang2016multi} & 85.19 \\
        \textbf{Our Model} & \textbf{85.61} \\
    \end{tabular}
    \centering
    \parbox{6.2cm}{\caption{Comparison with other works on CoNLL 2002 dataset (Dutch, NER).} \label{table_results_conll2002_ned}}
\end{table}

\begin{table}[ht]
    \small
    \vspace{0.5cm}
    \fontfamily{cmss}\selectfont
    \begin{tabular}{p{4.2cm}|>{\centering\arraybackslash}p{1.2cm}}
        Model & F$_{1}$ Score \\
        \hline
        \newcite{collobert2011natural} & 89.59 \\
        \newcite{huang2015bidirectional} & 90.10 \\
        \newcite{lample2016neural} & 90.94 \\
        \textbf{Our Model} & \textbf{91.11} \\
        \newcite{luo2015joint} & 91.20 \\
        \newcite{yang2016multi} & 91.20 \\
        \newcite{ma2016end} & 91.21 \\
        \newcite{chiu2016named} & 91.62 \\
        \newcite{peters2017semi} & 91.93
    \end{tabular}
    \centering
    \parbox{6.2cm}{\caption{Comparison with other works on CoNLL 2003 dataset (English, NER).} \label{table_results_conll2003}}
\end{table}

\section{Results} \label{results}

We present the results of our experiments in two different contexts. Table \ref{table_results_ablation} shows the performance of different model configurations gauged in ablation studies. Tables \ref{table_results_conll2000}, \ref{table_results_conll2002_esp}, \ref{table_results_conll2002_ned}, \ref{table_results_conll2003}, \ref{table_results_conll2012}, \ref{table_results_germeval2014}, \ref{table_results_wsjptb}, and \ref{table_results_tiger} juxtapose our results on each dataset with those of other works reporting their results on the same dataset. When citing results of other works, we indicate the best performance reported in the corresponding publication (independent of the methodology used). Each of our scores reported in Tables \ref{table_results_ablation}-\ref{table_results_tiger} was achieved by a trained model on the dataset's official test set. The scores were verified using CoNLL 2000 evaluation script\footnote{\url{https://www.clips.uantwerpen.be/conll2000/chunking/conlleval}}.

\begin{table}[ht]
    \small
    \fontfamily{cmss}\selectfont
    \begin{tabular}{p{4.2cm}|>{\centering\arraybackslash}p{1.2cm}}
        Model & F$_{1}$ Score \\
        \hline
        \newcite{ratinov2009design} & 83.45\tablefootnote{The result is taken from \newcite{durrett2014joint}.} \\
        \newcite{durrett2014joint} & 84.04 \\
        \newcite{chiu2016named} & 86.28 \\
        \newcite{strubell2017fast} & 86.99 \\
        \textbf{Our Model} & \textbf{87.84} \\
    \end{tabular}
    \centering
    \parbox{6.2cm}{\caption{Comparison with other works on CoNLL 2012 dataset (English, NER).} \label{table_results_conll2012}}
\end{table}

\begin{table}[ht]
    \small
    \vspace{0.5cm}
    \fontfamily{cmss}\selectfont
    \begin{tabular}{p{4.2cm}|>{\centering\arraybackslash}p{1.2cm}}
        Model & F$_{1}$ Score \\
        \hline
        \newcite{schuller2014mostner} & 73.24 \\
        \newcite{reimers2014germeval} & 76.91 \\
        \newcite{agerri2016robust} & 78.42 \\
        \newcite{hanig2014modular} & 79.08 \\
        \textbf{Our Model} & \textbf{79.21} \\
    \end{tabular}
    \centering
    \parbox{6.2cm}{\caption{Comparison with other works on GermEval 2014 dataset (German, NER).} \label{table_results_germeval2014}}
\end{table}

\begin{table}[ht]
    \small
    \fontfamily{cmss}\selectfont
    \begin{tabular}{p{4.2cm}|>{\centering\arraybackslash}p{1.2cm}}
        Model & Accuracy \\
        \hline
        \newcite{toutanova2003feature} & 97.24 \\
        \newcite{collobert2011natural} & 97.29 \\
        \newcite{santos2014learning} & 97.32 \\
        \newcite{sun2014structure} & 97.36 \\
        \textbf{Our Model} & \textbf{97.45} \\
        \newcite{ma2016end} & 97.55 \\
        \newcite{huang2015bidirectional} & 97.55 \\
        \newcite{yang2016multi} & 97.55 \\
        \newcite{ling2015finding} & 97.78 \\
    \end{tabular}
    \centering
    \parbox{6.2cm}{\caption{Comparison with other works on WSJ/PTB dataset (English, POS-tag.).} \label{table_results_wsjptb}}
\end{table}

\begin{table}[ht]
    \small
    \vspace{0.5cm}
    \fontfamily{cmss}\selectfont
    \begin{tabular}{p{4.2cm}|>{\centering\arraybackslash}p{1.2cm}}
        Model & Accuracy \\
        \hline
        \newcite{labeau2015non} & 97.14 \\
        \newcite{muller2013efficient} & 97.44 \\
        \newcite{nguyen2016robust} & 97.46 \\
        \newcite{muller2015robust} & 97.73 \\
        \newcite{ling2015finding} & 98.08\tablefootnote{The result is not directly comparable to ours, as the authors used different train/dev/test split of Tiger Corpus.} \\
        \textbf{Our Model} & \textbf{98.40} \\
    \end{tabular}
    \centering
    \parbox{6.2cm}{\caption{Comparison with other works on Tiger Corpus dataset (German, POS-tag.).} \label{table_results_tiger}}
\end{table}

The ablation studies examined different partial configurations of the full model described in Section \ref{model} evaluated on each of the eight datasets. The configurations were obtained by altering the contents of joint embeddings and omitting the CRF layer. Joint embeddings were set to only word, only byte, or both word and byte embeddings. For each of these three settings, the CRF layer was included or excluded, amounting to six configurations in total.

The results of the ablation studies in Table \ref{table_results_ablation} show that, on every dataset, word and byte embeddings used jointly substantially outperformed any one of them used individually. This emphasizes the importance of using both embedding types, bringing in both semantic and morphological information about the words, for solving the general sequence labeling task. The role of the CRF layer proved to be crucial for all NER and Chunking, but not POS-tagging tasks. Configurations with and without CRF layer, given the same word representation, demonstrated very similar results on both English and German POS-tagging datasets. This supports the conjecture that a CRF layer can provide substantial incremental benefit, when used for solving a structured prediction task (e.g., NER and Chunking). It is also worth mentioning that on five out of eight datasets, using byte embeddings alone yielded better results than using word embeddings alone, both with and without CRF layer. This may be caused by the fact that word embeddings are external (and not necessarily related) to the data, while byte embeddings are always learned from the data itself.

Comparison of our results with those of other works, shown in Tables \ref{table_results_conll2000} through \ref{table_results_tiger}, manifests that our model has achieved state-of-the-art performance on four out of eight datasets: namely, \mbox{85.61 F$_{1}$} on \mbox{CoNLL 2002} (Dutch NER), \mbox{87.48 F$_{1}$} on \mbox{CoNLL 2012} (English NER), \mbox{79.21 F$_{1}$} on \mbox{GermEval 2014} (German NER), and \mbox{98.40\%} on \mbox{Tiger Corpus} (German POS-tagging). We consider the scores obtained on the remaining four datasets to be competitive.

\section{Related Work} \label{related}

Arguably, the modern era of deep neural network-based NLP, in particular that of sequence labeling, has emerged with the work of \newcite{collobert2011natural}. Using pre-trained word and feature embeddings as word-level inputs, the authors have applied a CNN with "max over time" pooling and downstream MLP. As an objective, they  maximized "sentence-level log-likelihood", similar to the CRF-based approach from Section \ref{crf_layer}.

Since then, many more sophisticated neural models were proposed. Combining a Bi-LSTM network with a CRF layer to model sequence of words in a sentence was first introduced by \newcite{huang2015bidirectional}. The authors have employed substantial amount of engineered features, as well as external gazetteers. \newcite{ling2015finding} have proposed extracting morphological information from words using a character-level Bi-LSTM. The authors have combined the character-level embeddings with proprietary word embeddings (trained in-house) to obtain current stat-of-the-art result on the WSJ/PTB English POS-tagging dataset. 

\newcite{dos2015boosting} have augmented the model of \newcite{collobert2011natural} with a CNN-based character-level feature extractor network ("CharWNN") and applied the resulting model for NER. \newcite{chiu2016named} have combined CharWNN with a Bi-LSTM on word level. A similar approach was proposed by \newcite{ma2016end}. However, \newcite{ma2016end} did not use any feature engineering or external lexicons, as opposed to \newcite{chiu2016named}, and still obtained respectably high performance. 

One of the two models proposed by \newcite{lample2016neural} is quite similar to ours, except that the authors have assumed a fixed set of characters for each language, whereas we turn to bytes as a universal medium of encoding character-level information. Also, \newcite{lample2016neural} have pre-trained their own word embeddings, which turned out to have a crucial impact on their results.

An interesting approach of applying cross-lingual multi-task learning to sequence labeling problem was introduced in the work of \newcite{yang2016multi}. The authors have used hierarchical bi-directional GRU (on character and word levels) and optimized a modified version of a CRF objective function. Their model, in conjunction with the  applied multi-task learning framework, has allowed the authors to obtain state-of-the-art results on multiple datasets.

\section{Conclusion} \label{conclusion}

We evaluate the performance of a single general-purpose neural sequence labeling model, assuming unavailability of domain expertise and scarcity of informational and computational resources. The work explores the frontiers of what may be achieved with a generic and resource-efficient sequence labeling framework applied to a diverse set of NLP tasks and languages.

For this purpose, we've applied the model (\mbox{Section \ref{model}}) and the end-to-end training methodology (\mbox{Section \ref{training}}) to eight benchmark datasets (\mbox{Section \ref{datasets}}), covering four languages and three tasks. The obtained results have convinced us that, with a model of sufficient learning capacity, it is indeed possible to achieve competitive sequence labeling performance without the burden of delving into specificities of each particular task and language, and summoning additional resources.

For the future work, we envision the integration of multi-task learning techniques (e.g., those used by \newcite{yang2016multi}) into the proposed approach. We suppose that this may improve the current results without compromising our general applicability and practicality assumptions.

\bibliography{sequence_labeling}
\bibliographystyle{acl_natbib_nourl}

% \appendix

\end{document}